%% file: iclr2027_conference.tex
\documentclass{article} 
\usepackage{iclr2027_conference,times}

\input{math_commands.tex}

\usepackage{hyperref}
\usepackage{url}
\usepackage{multirow}
\usepackage{graphicx}
\usepackage{booktabs}
\usepackage{tabularx}
\usepackage{array}
\usepackage{xcolor}
\usepackage{colortbl}
\usepackage{soul}
\usepackage{siunitx}
\usepackage{enumitem}
\usepackage{wrapfig}
\usepackage{hhline}

\definecolor{mygray}{gray}{0.9}
\definecolor{my_blue}{HTML}{d3eaf2}
\newcolumntype{Y}{>{\centering\arraybackslash}X}
\newcolumntype{L}[1]{>{\raggedright\arraybackslash}m{#1}}
\newcolumntype{C}[1]{>{\centering\arraybackslash}m{#1}}

\usepackage{color}
\definecolor{citecolor}{RGB}{66,168,235}
\definecolor{linkcolor}{RGB}{255,0,0}

\hypersetup{colorlinks=true,citecolor=citecolor,linkcolor=linkcolor}

\title{LISA: Likelihood Score Alignment for Visual-condition Controllable Generation}

\author{Yanghao Wang$^{1}${\quad}Hongxu Chen$^{1}${\quad}Jiazhen Liu$^{1}${\quad} Zhenqi He$^{1}$\\
\textbf{Rui Liu$^{2}${\quad}Zhen Wang$^{1}${\quad}Long Chen$^{1}$}\thanks{Corresponding author. \hfill \textbf{Project page:} \href{https://github.com/HKUST-LongGroup/LISA}{https://github.com/HKUST-LongGroup/
LISA}}
\\
$^1$The Hong Kong University of Science and Technology\quad $^2$Huawei Research 
\\
\small \texttt{\{ywangtg,hchenej,jliugj,zheci\}@connect.ust.hk} \\ 
\small \texttt{ruiliu011@gmail.com\quad\{zhenwang,longchen\}@ust.hk} 
}

\newcommand{\ie}{\textit{i.e.}}
\newcommand{\eg}{\textit{e.g.}}
\newcommand{\cf}{\textit{c.f.}}

\newcommand{\x}{\bm{x}}
\newcommand{\grad}{\nabla_{\bm{x}_t}\log p_t}
\newcommand{\xt}{\bm{x}_t}
\newcommand{\xT}{\bm{x}_T}
\newcommand{\xzero}{\bm{x}_0}
\newcommand{\cond}{\bm{c}}

\iclrfinalcopy 
\begin{document}

\maketitle

\input{./sections/0_abstract}
\input{./sections/1_intro}
\input{./sections/2_preliminaries}
\input{./sections/3_method}
\input{./sections/4_experiments}
\input{./sections/5_related_works}
\input{./sections/6_conclusion}

\bibliography{iclr2027_conference}
\bibliographystyle{iclr2027_conference}

\input{sections/7_appendix}

\end{document}

%% file: math_commands.tex
\usepackage{amsmath,amsfonts,bm}

\def\eqref#1{equation~\ref{#1}}

\def\1{\bm{1}}

\DeclareMathAlphabet{\mathsfit}{\encodingdefault}{\sfdefault}{m}{sl}
\SetMathAlphabet{\mathsfit}{bold}{\encodingdefault}{\sfdefault}{bx}{n}

\DeclareMathOperator*{\argmin}{arg\,min}

%% file: sections/0_abstract.tex
\begin{abstract}
The prevalent \emph{dual-branch paradigm}, \ie, training a side network to encode visual conditions and fusing its intermediate-layer features to a frozen pretrained main network, has shown remarkable success in visual-condition controllable generation. 
Despite its widespread adoption, the role of the side branch and its training efficiency remain underexplored.
In this paper, we first revisit this mainstream paradigm through the lens of score-based generative modeling: 1) The main network preserves visual perceptual quality by providing a prior unconditional score. 2) The side network steers conditional control by implicitly contributing a likelihood score.
Guided by this perspective, we propose \emph{LIkelihood Score Alignment (\textbf{LISA})}, an effective regularization method that explicitly aligns the intermediate features of the side network with an approximated likelihood score. Specifically, we first hook features from a designated layer of the side network and project them into the score latent space by a lightweight decoder. 
Then, we construct an approximated likelihood score target and calculate the distance between the decoder's output and this target as an additional regularization loss.
Finally, we jointly optimize the side network and decoder with both standard diffusion loss and our regularization loss. Experiments across various image/video tasks, architectures, and diffusion/flow models demonstrate that LISA can not only consistently accelerate the training convergence and improve final synthetic results, but also encourage the side network's features to be more disentangled for conditional modeling
with negligible additional training cost and zero extra inference cost.
\end{abstract}
\input{./figures/fig_intro}

%% file: figures/fig_intro.tex
\begin{figure}[h]
    \centering
    \includegraphics[width=1\linewidth]{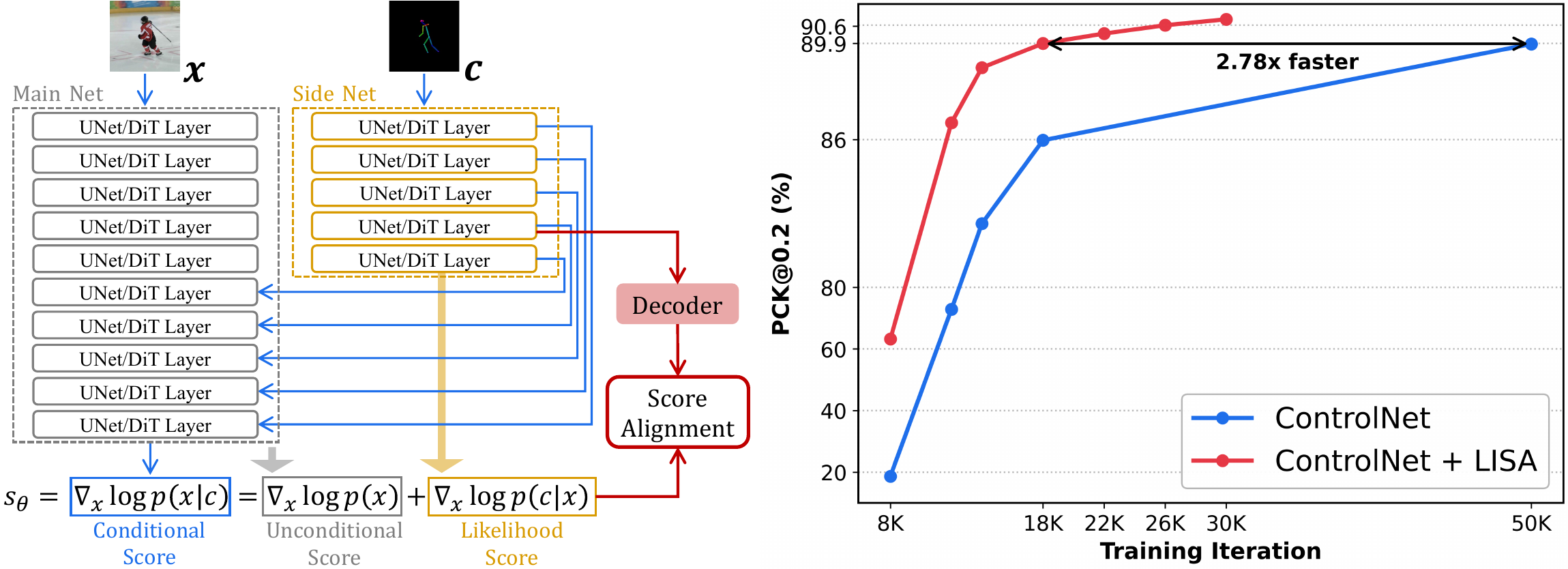}
    \vspace{-2em}
    \caption{\textbf{Likelihood score alignment (LISA) can improve training convergence and synthetic quality}. Our framework, LISA, explicitly decomposes roles within the dual-branch paradigm: the main network and side network are responsible for the \emph{unconditional} and \emph{likelihood score}, respectively. By aligning a certain feature of the side network with an approximated likelihood score via a lightweight decoder, LISA can achieve $>2.78\times$ faster convergence (\eg, as in ControlNet).}
    \label{fig:intro}
    \vspace{-1em}
\end{figure}

%% file: sections/1_intro.tex
\section{Introduction}
\label{sec:1}

Recent advances in diffusion~\citep{ho2020denoising,song2020score} and flow matching models~\citep{liu2022flow,lipman2022flow} show remarkable visual generation capability. In particular, unconditional and text-conditioned generation tasks have been addressed quite well by billion-parameter models~\citep{esser2024scaling,flux-2-2025,wan2025wan,hacohen2026ltx} trained on large-scale and easy-to-collect training data. However, the increasing application requirements introduce a more challenging scenario~\citep{batzolis2021conditional}: \textit{Visual-condition Controllable Generation}, \ie, integrating visual-modality conditions, especially spatial conditions (\eg, pose, segmentation, and depth maps) for more fine-grained, structurally controllable image and video generation.

To achieve this goal, prior studies~\citep{zhang2023adding,mou2024t2i,strawberry_svdtemporalcontrolnet_2023} resort to a \textbf{dual-branch paradigm} (\cf, Figure~\ref{fig:intro}): 1) Freezing a pretrained diffusion (or flow matching) model as the \emph{main network} backbone. 2) Training a condition encoder as the \emph{side network} to adopt the condition input and output a set of intermediate features. 3) Integrating these features into the main network's original forward process to achieve conditional control. 
Under this paradigm, the follow-up study~\citep{xie2026divcontrol} extends representation alignment technology~\citep{yu2024representation} to controllable generation, \ie, aligning model features with a pretrained semantic encoder as an additional regularization to improve training efficiency. However, the dependence on external encoders bounds their performance to the chosen encoder.

In this paper, we revisit this paradigm through the lens of score-based generative modeling: each branch network implicitly plays decomposed roles, and the feature-level integration mechanism essentially tries to induce an augmented result: 1) The frozen main network does not adopt the condition input, \ie, it is mainly responsible for providing an \textit{unconditional score} $\grad (\xt)$ to guarantee general perceptual quality. 2) The trainable side network encodes the condition $\cond$ and learns to bridge the gap between the \textit{conditional score} $\grad (\xt|\cond)$ and the \textit{unconditional score} $\grad (\xt)$, \ie, achieving conditional control. 3) According to Bayes' rule, this residual gap is the \textit{likelihood score}, \ie, $\grad (\cond|\xt) = \grad (\xt|\cond) - \grad (\xt)$.

\textbf{Our approach.} Based on the above analyses, we argue that the main challenge in the dual-branch paradigm lies in training the side network, which learns to provide the control signal and implicitly corresponds to a likelihood score. Motivated by this, we propose \textit{\ul{LI}kelihood \ul{S}core \ul{A}lignment} (LISA), an effective regularization technique that explicitly aligns the side network with an approximated likelihood score.
Since the frozen main network is naturally an unconditional score predictor and the paired training data can provide the closed-form conditional score, we can estimate a likelihood score by calculating the difference between them. By aligning the intermediate features of the side network with this approximated score alongside standard generative training, LISA introduces an efficient prior supervision. This explicit constraint acts as a regularization loss, significantly accelerating convergence and improving overall synthesis performance (\cf, Figure~\ref{fig:intro}).
Meanwhile, such regularization can encourage the side network's features to be more disentangled for conditional modeling, thereby naturally demonstrating better compositional control.
It is worth noting that, compared with existing representation alignment, our LISA does not require external semantic encoders and achieves comparable improvements in training convergence and synthesis quality.

Specifically, during the standard generative training, we first hook features from a designated layer of the side network and feed them into a lightweight trainable decoder (usually around $0.1\%$ of the size of the side network). This decoder only contains a few layers (\eg, convolution, activation, and upsampling layers) to transform the intermediate features of the side network into a latent score space. Then, we calculate the distance between the decoder output and the likelihood score as a regularization loss, which is added to the diffusion loss as the final optimization objective. Notably, the additional training cost introduced by such regularization is almost negligible. During inference, we directly drop the decoder and use the trained side network for final conditional generation.

We evaluate the effectiveness of our LISA across various image and video conditions, including pose maps, depth maps, low-resolution images, segmentation maps, and pose videos. All experimental results have consistently indicated that LISA can significantly accelerate training convergence and bootstrap to a better synthetic quality (both perceptual quality and condition fidelity). Further evaluations, such as architecture-agnostic generalization and more challenging compositional controls, have verified that LISA is an effective and extensible solution for visual-condition controllable generation. 
In summary, our main contributions are as follows:
\vspace{-1em}
\begin{itemize}[leftmargin=*]
\itemsep-0.2em
    \item We analyzed the mainstream dual-branch paradigm for visual-condition generation from a novel perspective, revealing that the side network lacks explicit regularization for its intended role.
    \item Based on the role decomposition, we proposed an effective likelihood alignment method, LISA, which regularizes the side network's intermediate output with an approximated likelihood score.
    \item Extensive experiments across diffusion models, network architectures, and tasks have demonstrated significant and consistent gains in both training convergence and synthesis quality.
\end{itemize}

%% file: sections/2_preliminaries.tex
\section{PRELIMINARIES}
\label{sec:2}
We present a brief overview of diffusion and flow matching models via the unified perspective of the stochastic process~\citep{jolicoeur2021gotta} and score matching~\citep{song2019generative}.

The diffusion/flow models aim to capture the target data distribution $p_{0}$ by learning a transport process from a prior distribution $p_{T}$ (\eg, a Gaussian distribution) to $p_{0}$. To achieve that, a forward diffusion process from $p_{0}$ to $p_{T}$ can be described with such a stochastic differential equation (SDE):
\begin{equation}
\label{eq:1}
    \mathrm{d}\x = \bm{f}(\xt,t)\mathrm{d}t + g(t)\mathrm{d}\bm{w},
\end{equation}
where $t\in[0,T]$ is the time-index, $\xzero \sim p_{0}$, $\bm{w}$ is a Brownian motion, $\bm{f}$ and $g$ are the drift and diffusion coefficient. 
Meanwhile, the marginal distribution of $\xt$ determined by this SDE is $p_t(\xt)$.
Then, we have a reversed SDE to describe the reversed process ( $p_{T}$ $\rightarrow$ $p_{0}$) of the forward process:
\begin{equation}
\label{eq:2}
    \mathrm{d}\x = \left[\bm{f}(\xt,t)-g^2(t)\grad(\xt) \right]\mathrm{d}t + g(t)\mathrm{d}\overline{\bm{w}},
\end{equation}
where $\overline{\bm{w}}$ is the reversed Brownian motion. According to the Fokker–Planck equation~\citep{maoutsa2020interacting} (F-P Equation), we can convert this reversed SDE into an ordinary differential equation (ODE) that has the same marginal distribution $p_t(\xt)$:
\begin{equation}
\label{eq:3}
    \mathrm{d}\x = \left[\bm{f}(\xt,t)-\frac{1}{2}g^2(t)\grad(\xt)\right]\mathrm{d}t. 
\end{equation}
We can sample synthetic samples by solving the SDE in Eq.~(\ref{eq:2}) or the ODE in Eq.~(\ref{eq:3}). However, $\grad(\xt)$ (also known as the unconditional score) is not known. Thus, a parameterized network $s_{\theta}(\cdot)$ with parameters $\theta$ can be trained to predict it. The optimization target is:
\begin{equation}
\label{eq:4}
    \mathop{\mathrm{min}}_{\theta} \mathbb{E}_{\xzero,t,\xt} \left[||s_\theta(\xt,t)-\grad(\xt|\xzero)||_2^2 \right],
\end{equation}
where $\xzero\sim p_0$, $t \in [0,T]$ and $\xt\sim p_t(\xt|\xzero)$. $\grad(\xt|\xzero)$ is tractable since the conditional distribution is defined by the forward SDE in Eq.~(\ref{eq:1}), and it has closed-form solutions. After training, $s_{\theta}$ can be used as a score predictor to replace $\grad(\xt)$ for solving Eq.~(\ref{eq:2}) and Eq.~(\ref{eq:3}). 

%% file: sections/3_method.tex
\section{LISA: Likelihood Score Alignment}
\label{sec:3}
\input{./figures/fig_method}

\textbf{Problem Formulation}. Assume there exists an underlying joint distribution $(\xzero,\cond) \sim p$, which describes the joint probability measurement between visual-modality conditions $\cond$ and corresponding clean samples $\xzero$. For the conditional visual generation task, our objective is to use given $(\xzero, \cond)$ pairs to train a conditional score predictor $s(\xt,\cond,t)$. In the inference stage, we use $s(\xt,\cond,t)$ to transport a random noise $\xT$ to a clean sample $\xzero$ conditioned on the given condition $\cond$\footnote{For clarity, we omit text prompts in notations, and when needed they can be regarded as extra conditions.}.

\subsection{Score Decomposition of the Dual-Branch Paradigm}
\label{sec:score_decomposition}

\textbf{Standard Conditional Diffusion}.
Given the joint data distribution $\xzero,\cond$, the noisy conditional distribution $p_t(\xt|\cond)$ induced
by the forward SDE of Eq.~(\ref{eq:1}) is:
\begin{equation}
    p_t(\xt|\cond)=\int p(\xzero|\cond)p_t(\xt|\xzero)\mathrm{d}\xzero .
\end{equation}
Taking the gradient with respect to $\xt$, the conditional score can be written as:
\begin{align}
    \nabla_{\xt}\log p_t(\xt|\cond)
    =
    \frac{\nabla_{\xt}p_t(\xt|\cond)}{p_t(\xt|\cond)} \nonumber
    &=
    \int
    p_t(\xzero|\xt,\cond)
    \nabla_{\xt}\log p_t(\xt|\xzero)
    \mathrm{d}\xzero \nonumber\\
    &=
    \mathbb{E}_{\xzero |\xt,c}\left[
        \grad (\xt|\xzero)
    \right].
\label{eq:conditional_score_identity}
\end{align}
Eq.~(\ref{eq:conditional_score_identity}) shows that the conditional score
$\nabla_{\xt}\log p_t(\xt|\cond)$ is the expectation of the clean-sample score
$\grad(\xt|\xzero)$. Therefore, although $\grad(\xt|\xzero)$ is conditioned on the clean sample $\xzero$ rather than directly on $\cond$, it provides an unbiased supervision for learning the conditional score.

Thus, for the dual-branch paradigm, a conditional score predictor for visual-condition generation is trained with the standard denoising score matching objective:
\begin{equation}
\label{eq:l_main}
    \mathcal{L}_{\mathrm{main}}
    =
    \mathbb{E}_{\xzero, \cond, t,\xt}
    \left[
        \left\|
            s_{\theta,\phi}(\xt,\cond,t)-\grad (\xt|\xzero)
        \right\|_2^2
    \right],
\end{equation}
where $\theta$ and $\phi$ denote the parameters of the frozen pretrained main network and the trainable side network.
The optimal solution of Eq.~(\ref{eq:l_main}) is $s_{\theta,\phi}^{*}(\xt,\cond,t)=\nabla_{\xt}\log p_t(\xt|\cond)$.

\textbf{Dual-branch Decomposition}.
We now decompose this conditional score with Bayes' rule:
\begin{align}
    \nabla_{\xt}\log p_t(\xt|\cond)
    &=
    \nabla_{\xt}\log p_t(\xt)
    +
    \nabla_{\xt}\log p_t(\cond|\xt),
\label{eq:bayes_score_decomposition}
\end{align}
where the term $\nabla_{\xt}\log p_t(\cond)$ disappears because $\log p_t(\cond)$ is independent of $\xt$. Eq.~(\ref{eq:bayes_score_decomposition}) indicates that conditional generation can be interpreted as the combination of two scores: the unconditional score $\nabla_{\xt}\log p_t(\xt)$ and the likelihood score $\nabla_{\xt}\log p_t(\cond|\xt)$.

This decomposition naturally matches the dual-branch paradigm (\cf, Figure~\ref{fig:method}). We denote the side network by $r_{\phi}^i$ and its intermediate features by
$\{r_{\phi}^{i}(\xt,\cond,t)\}_{i=1}^{L}$, where $L$ is the number of selected side features. The full conditional score predictor can be written as:
\begin{equation}
    s_{\theta,\phi}(\xt,\cond,t)
    =
    \mathcal{S}_{\theta}
    \left(
        \xt,t;\{r_{\phi}^{i}(\xt,\cond,t)\}_{i=1}^{L}
    \right),
\end{equation}
where disabling the side features gives the frozen main-network prediction $s_{\theta}(\xt,t)=\mathcal{S}_{\theta}(\xt,t;\emptyset)$. Since $s_{\theta}$ has been pretrained to approximate $\nabla_{\xt}\log p_t(\xt)$, the trainable side network is implicitly responsible for supplying the residual correction from the unconditional score to the conditional score, which is called the likelihood score:
\begin{equation}
    s_{\theta,\phi}(\xt,\cond,t)-s_{\theta}(\xt,t)
    \approx
    \nabla_{\xt}\log p_t(\cond|\xt).
\end{equation}
However, the standard objective in Eq.~(\ref{eq:l_main}) supervises only the final prediction, leaving this likelihood-score role of the side network implicit. This motivates us to explicitly align side-network features with an approximated likelihood score.


\subsection{Alignment with Approximated Likelihood Score}
\label{sec:lisa_alignment}

\textbf{Approximated Likelihood Score Construction}.
To explicitly supervise the side network during training, we first need to obtain a likelihood-score target.
According to Bayes' rule, the likelihood score can be written as the difference between the conditional and unconditional scores:
\begin{align}
    \nabla_{\xt}\log p_t(\cond|\xt)
    &=
    \nabla_{\xt}\log p_t(\xt|\cond)
    -
    \nabla_{\xt}\log p_t(\xt).
    \label{eq:likelihood_score_decomposition}
\end{align}
Although the conditional score $\nabla_{\xt}\log p_t(\xt|\cond)$ is intractable, we can use the denoising target $\grad(\xt|\xzero)$ to provide a single-sample supervision signal whose expectation equals the conditional score (\cf, Eq.~(\ref{eq:conditional_score_identity})).

Meanwhile, the pretrained main network is naturally an unconditional score predictor. To this end, we additionally forward the main network without any condition injection to obtain
$s_{\theta}(\xt,t)\approx\nabla_{\xt}\log p_t(\xt)$ (\cf, Figure~\ref{fig:method}).
Substituting two estimates into Eq.~(\ref{eq:likelihood_score_decomposition}):
\begin{align}
    \nabla_{\xt}\log p_t(\cond|\xt)
    &\approx
    \mathbb{E}_{\xzero |\xt,c}
    \left[
        \grad(\xt|\xzero)
        -
        s_{\theta}(\xt,t)
    \right].
\end{align}
This motivates the following sample-wise approximated likelihood score:
\begin{equation}
\label{eq:approximated_likelihood}
    \widehat{\ell}_t(\cond|\xt)
    =
    \grad(\xt|\xzero)
    -
    s_{\theta}(\xt,t).
\end{equation}
Therefore, Eq.~(\ref{eq:approximated_likelihood}) provides a practical and efficient target for likelihood-score alignment.

\textbf{Likelihood Score Alignment}.
To align the side network with this target, we select a certain feature $r_{\phi}^{k}(\xt,\cond,t)$ (the $k$-th layer's output) from the side network before integrating. We feed it into a lightweight decoder $\mathcal{D}_{\psi}$ composed of convolution, activation, and upsampling layers:
\begin{equation}
    \widetilde{\ell}_{\psi}^{k}(\cond|\xt)
    =
    \mathcal{D}_{\psi}
    \left(
        r_{\phi}^{k}(\xt,\cond,t),t,\xt
    \right).
\end{equation}
The decoder maps the selected side feature into the same latent score space as the diffusion target. We then impose the LISA regularization loss:
\begin{equation}
\label{eq:lisa_loss}
    \mathcal{L}_{\mathrm{LISA}}
    =
    \mathbb{E}_{\xzero,\cond,t,\xt}
    \left[
        \left\|
            \widetilde{\ell}_{\psi}^{k}(\cond|\xt)
            -
            \operatorname{sg}
            \left[
                \widehat{\ell}_t(\cond|\xt)
            \right]
        \right\|_2^2
    \right],
\end{equation}
where $\operatorname{sg}[\cdot]$ denotes stop-gradient operation. Combining the standard diffusion loss, the final objective is
\begin{equation}
\label{eq:total_loss}
    \phi^*,\psi^*=\mathop{\argmin}_{\phi,\psi}(
    \mathcal{L}_{\mathrm{main}}
    +
    \lambda
    \mathcal{L}_{\mathrm{LISA}}),
\end{equation}
where $\lambda$ controls the strength of likelihood-score alignment. During training, the parameters of the main network are frozen. We optimize the side network and the lightweight decoder jointly. The standard loss $\mathcal{L}_{\mathrm{main}}$ supervises the final conditional prediction, while $\mathcal{L}_{\mathrm{LISA}}$ provides a direct auxiliary gradient to the side network. Therefore, LISA makes the side network learn its intended likelihood-score role more explicitly and efficiently. During inference, the auxiliary decoder is discarded, and the trained side network is used exactly in the same way as the original architecture.

%% file: figures/fig_method.tex
\begin{figure}[t]
    \centering
    \includegraphics[width=1\linewidth]{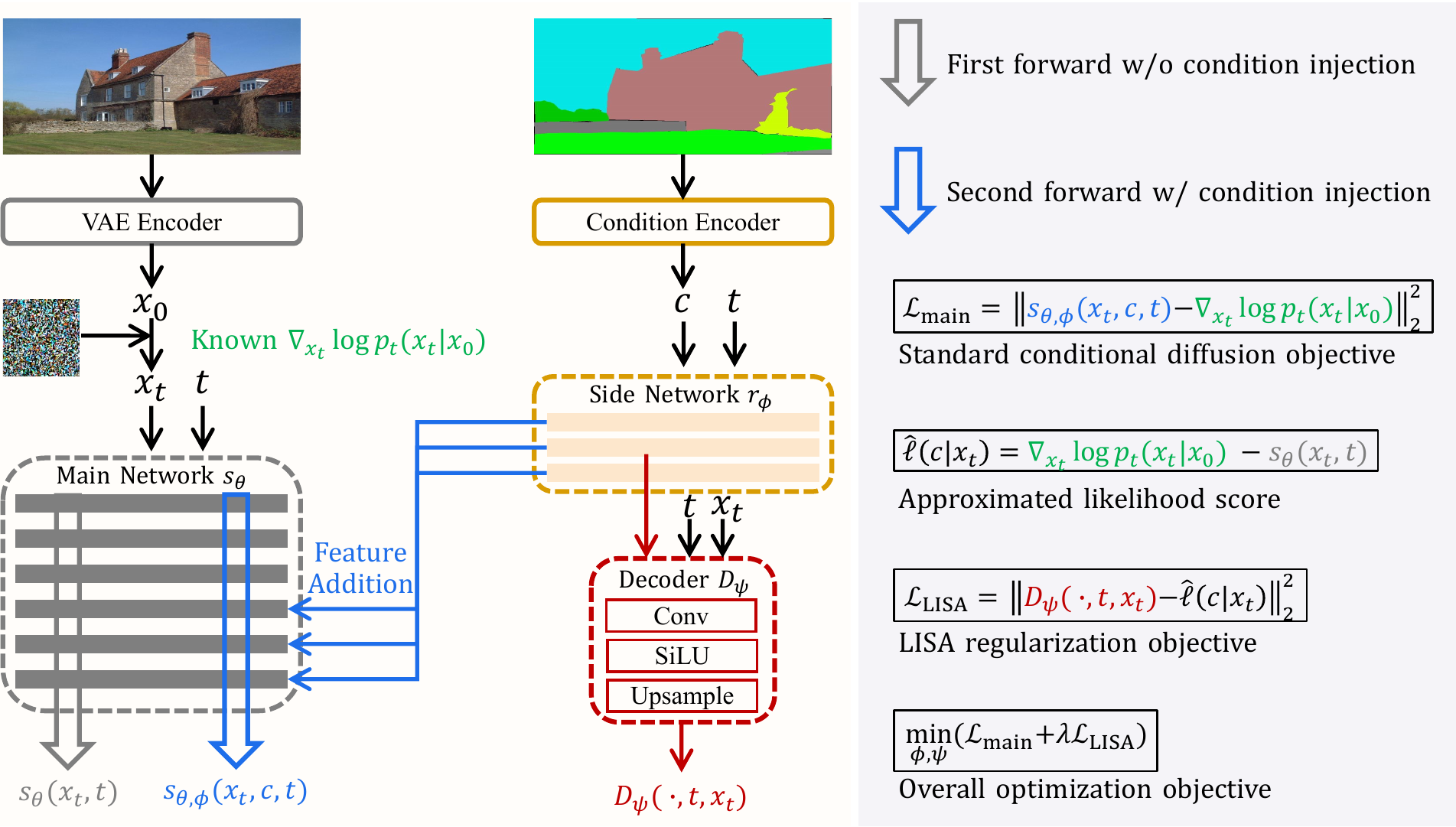}
    \caption{\textbf{The framework of LISA}. The first forward w/o condition injection provides the unconditional score $s_{\theta}(\xt,t)$. By subtracting it from the known $\nabla_{\xt}\log p_t(\xt|x_0)$, we can construct an approximated likelihood score $\hat{\ell}_{t}(\xt,c)$. In the second forward w/ condition injection, we align the features of the side network with the $\hat{\ell}_{t}(\xt,c)$ via a decoder as an extra regularization objective.}
    \label{fig:method}
\end{figure}

%% file: sections/4_experiments.tex
\section{Experiments}
\label{sec:4}

\subsection{Main Results}
\label{sec:4.1}
To verify the effectiveness, we compared LISA with the vanilla representative dual-branch baselines: T2I-Adapter~\citep{mou2024t2i}, ControlNet~\citep{zhang2023adding}, and ControlNet+REPA~\citep{xie2026divcontrol} across four types of conditional image generation tasks: Pose~\citep{cao2017realtime}, ADE20K Segmentation~\citep{zhou2017scene}, Depth~\citep{ranftl2020towards} maps and low-resolution image conditional generation. We took SDXL-1.0~\citep{podell2024sdxl} and Stable Diffusion 2.1~\citep{rombach2022high} as the pretrained diffusion model for T2I-Adapter and ControlNet, respectively. 
For a fair comparison, we used AdamW~\citep{loshchilov2017decoupled} optimizer with a $1e-5$ learning rate and maintained all the hyperparameters the same as the baseline. For the REPA implementation, we followed DIvControl~\citep{xie2026divcontrol}: align the feature (same layer as our LISA) with DINOv2-B~\citep{oquab2023dinov2} using a regularization weight of $0.05$.

\input{./tables/table1}
\noindent\textbf{Metrics}.
For four image-conditioned generation tasks, we adopted the Frechet Inception Distance (\textit{FID})~\citep{heusel2017gans}, which quantifies the distributional similarity between synthetic and ground-truth images. 
Besides, for the pose-conditioned task, we used averaged CLIP similarity (\textit{CLIP})~\citep{radford2021learning} between given text prompts and generated images to quantify the text condition following performance, as well as Percentage of Correct Keypoints~\citep{yang2011articulated} at the threshold of $0.2$ (\textit{PCK}) to quantify the pose condition following performance.
For the segmentation-conditioned task, we used \textit{CLIP} and mean Intersection over Union (\textit{mIoU})~\citep{everingham2010pascal} to quantify the segmentation condition following performance.
For the low-resolution-conditioned task, we used Peak Signal-to-Noise Ratio (\textit{PSNR}) and Learned Perceptual Image Patch Similarity (\textit{LPIPS})~\citep{zhang2018unreasonable}, which focus on image-level similarity between ground truth and generated images to quantify the effectiveness.
For the depth-conditioned task, we used \textit{CLIP} and Root Mean Square Error (\textit{RMSE})~\citep{eigen2014depth} to quantify the depth-conditioned following performance. 

\textbf{Quantitative Results}.
As shown in Table~\ref{tab:main}:
(1) LISA consistently improves the condition-following ability across all four image tasks. In the early training stage, LISA achieves substantial gains in structure-related metrics, \eg, improving \textit{PCK} from $19.38$ to $83.02$ for pose-conditioned image generation relative to ControlNet.
(2) LISA achieves better performance with fewer training iterations, demonstrating improved training efficiency. For example, in depth-conditioned image generation, LISA trained for only $4$K iterations obtains better \textit{FID}, \textit{CLIP}, and \textit{RMSE} than ControlNet trained for $10$K iterations.
(3) Compared with REPA, LISA achieved comparable performance without depending on any extra pretrained models.
Overall, these results indicate that LISA not only enhances the fidelity of synthesis but also accelerates convergence.

\input{./figures/fig_main}

\noindent\textbf{Qualitative Results}. We also gave qualitative comparisons in Figure~\ref{fig:main}.
We can see that LISA shows better condition-following capability across various settings, as well as a decent visual quality (more natural with fewer artifacts). For example, in the second pose-conditioned example, ControlNet generated an image with the person inverted front-to-back, while ours produced the correct pose.

\subsection{Ablation Study}
We ablated two main hyperparameters: the feature depth used for alignment and the weight $\lambda$ (\cf, Eq.~(\ref{eq:total_loss})). We used the pose-conditioned image generation task and ControlNet+LISA (with $18$K training iterations) as the default setting. Besides, we provided a computational overhead analysis.

\noindent\textbf{Alignment Depth}.
We studied the effect of the alignment depth while fixing $\lambda=0.2$.
As shown in Table~\ref{tab:ablation}, introducing the alignment module consistently improves the pose consistency measured by \textit{PCK} compared with the baseline.
Specifically, using alignment depths of $2$ and $8$ improves \textit{PCK} from $85.97\%$ to $88.03\%$ and $88.06\%$, respectively, while maintaining comparable \textit{FID} and \textit{CLIP}.
Among different depths, setting the depth to $5$ achieves the best \textit{PCK} of $89.90\%$.
A shallower alignment may be insufficient to fully capture structural correspondence, whereas an overly deep alignment does not bring further improvement and may introduce redundant constraints.
Therefore, we adopt an alignment depth of $5$ in our final implementations for all four conditional tasks.

\input{./tables/table2}
\noindent\textbf{Effect of $\lambda$}.
We further analyzed the influence of the loss weight $\lambda$ (\cf, Eq.~\ref{eq:total_loss}) with the alignment depth fixed to $5$.
When $\lambda=0.1$, the model obtains a lower \textit{PCK} of $86.19\%$, suggesting that a weak alignment constraint is insufficient to guide pose-consistent generation.
Increasing $\lambda$ to $0.5$ slightly improves FID to $56.34$, but the \textit{PCK} drops to $87.83\%$, indicating that an overly strong alignment constraint may hurt structural matching.
In contrast, $\lambda=0.2$ achieves the best overall balance, yielding the highest \textit{PCK} of $89.90\%$.
Thus, we set $\lambda=0.2$ as the default configuration.

\input{./tables/table3}
\noindent\textbf{Computational Overhead Analysis}.
We compared the computational cost on $8$ H20 GPUs between ControlNet and our LISA in Table~\ref{tab:overhead}. 
LISA introduces only a negligible number of additional parameters, increasing the model size from $364.2$M to $364.6$M, \ie, about 0.1\% extra parameters. 
Both methods require the same GPU memory consumption of $21$G, showing that the proposed alignment module does not increase the memory footprint. 
In terms of training time per iteration, LISA takes $2.3$s compared with $2.1$s for ControlNet, introducing only $0.2$s additional latency. As a highlight, during the inference stage, LISA directly drops the decoder, and thus the computational cost is completely the same as naive ControlNet.
These results indicate that LISA improves performance with minimal computational overhead, making it efficient and practical for deployment.

\subsection{Generalization Study}

\input{./tables/table4}
\input{./figures/fig_svd}
\noindent\textbf{Extend to Flow and DiT}. Since our main results are based on the U-Net~\citep{ronneberger2015u} along with the Variance-Preserving (VP) SDE, \ie, Stable Diffusion v2.1~\citep{rombach2022high}, we further test our effectiveness on Diffusion Transformer~\citep{peebles2023scalable} (DiT) along with Optimal Transport Flow Matching~\citep{liu2022flow,lipman2022flow} (OT-FM), \eg, Stable Diffusion v3-medium~\citep{esser2024scaling}. To this end, we conducted the same segmentation-conditioned experiments as in Section~\ref{sec:4.1} with the $8$-th layer output.

As shown in Table~\ref{tab:4}, we can see that at the early training stage, \ie, 1K iterations, LISA reduces FID from $32.08$ to $31.87$ and improves mIoU from $20.81\%$ to $22.64\%$, while maintaining a comparable CLIP score. When trained for $5$K iterations, LISA further improves all metrics, achieving lower FID, higher CLIP score, and higher mIoU compared with the ControlNet baseline. These results demonstrate that LISA is not limited to the U-Net architecture or VP-SDE formulation, but can also generalize well to diffusion transformers trained with flow matching objectives.

\noindent\textbf{Extend to Controllable Video Generation}. To further verify our generalization for controllable video generation, we also compared our LISA with SVD Temporal ControlNet (SVD Control)~\citep{strawberry_svdtemporalcontrolnet_2023} based on a pretrained image-to-video model, \ie, Stable Video Diffusion~\citep{blattmann2023stable} for the pose-guided video generation task on the UBC Fashion dataset~\citep{zablotskaia2019dwnet}. We used the same hyperparameters as in Section~\ref{sec:4.1}. There are four reported metrics: Frechet Video Distance~\citep{unterthiner2018towards} (\textit{FVD}), which can indicate distributional differences between the ground-truth videos and the synthetic videos, frame-level \textit{SSIM}~\citep{wang2004image}, \textit{LPIPS}, and \textit{PCK}. 

As shown in Table~\ref{tab:5}, our LISA also generalizes well to conditional video generation. At $5$K iterations, LISA significantly improves all metrics, \eg, reducing \textit{FVD} from $10.57$ to $7.85$ while increasing \textit{PCK} from $30.22\%$ to $57.00\%$. At $30$K iterations, LISA further maintains consistent gains over the ControlNet baseline across all metrics. These improvements demonstrate that LISA can effectively enhance both generation quality and condition controllability for video diffusion models with good generalization. We also provide the visualization case in Figure~\ref{fig:svd}.



\subsection{Bonus: Compositional-condition generation}
\input{./figures/fig_compose}
Since the alignment between the feature and the likelihood score can encourage the side network to model the condition more independently, the features under such regularization potentially should show more disentangled control performance.
To verify this, we further investigated whether LISA benefits the composition of multiple visual conditions. 
To fairly evaluate the compositional ability, we took independently trained single-condition side networks for pose and segmentation, where ControlNet and LISA show comparable performance under the corresponding single-condition settings. 
During inference, we directly composed the two conditions by summing their injected features at each corresponding layer, resulting in a pose-plus-segmentation conditioned generation setting. 

As shown in Figure~\ref{fig:compose}, LISA demonstrates stronger compositional generation ability than the naive ControlNet baseline with both quantitative and qualitative evidence.
These results suggest that the explicit role decomposition and likelihood-score alignment introduced by LISA help disentangle the main network and side networks, making the learned condition representations more composable and thus more suitable for multi-condition controllable generation as an extra bonus.

%% file: tables/table1.tex
\begin{table}[t]
\centering
\caption{Comparisons with dual-branch baselines across four conditional image generation tasks.}
\label{tab:main}
\setlength{\tabcolsep}{0pt}
\renewcommand{\arraystretch}{1}

\resizebox{0.9\columnwidth}{!}{%
\begin{tabularx}{1\columnwidth}{@{} l YYYY @{\hspace{1em}} YYYY @{}}
\hline\hline
\multicolumn{1}{c}{\multirow{2}{*}{Method}}
& \multicolumn{4}{c}{Pose Map} & \multicolumn{4}{c}{Segmentation Map} \\
\cmidrule(lr){2-5} \cmidrule(lr){6-9}
& Iter. & FID$_{\downarrow}$ & CLIP$_{\uparrow}$ & PCK(\%)$_{\uparrow}$
& Iter. & FID$_{\downarrow}$ & CLIP$_{\uparrow}$ & mIoU(\%)$_{\uparrow}$ \\
\midrule
T2I-Adapter
& 10K & 59.10 & 32.27 & 84.85 & 10K & 32.12& 31.67&29.07 \\
\cellcolor{my_blue}\small\textbf{+ LISA (ours)}
& \cellcolor{my_blue}10K & \cellcolor{my_blue}\textbf{58.12} & \cellcolor{my_blue}\textbf{32.30} & \cellcolor{my_blue}\textbf{85.94} & \cellcolor{my_blue}10K &\cellcolor{my_blue}\textbf{32.07} & \cellcolor{my_blue}\textbf{31.70} & \cellcolor{my_blue}\textbf{29.53}\\
\cmidrule[0.5pt](lr){2-5} \cmidrule[0.5pt](lr){6-9}
ControlNet  & 10K & 56.37 & \textbf{31.47} & 19.38
& 10K & 34.98 & \textbf{31.96} & 10.10 \\
\cellcolor{my_blue}\small\textbf{+ LISA (ours)}
& \cellcolor{my_blue}10K & \cellcolor{my_blue}\textbf{56.28} & \cellcolor{my_blue}31.39 & \cellcolor{my_blue}\textbf{83.02}
& \cellcolor{my_blue}10K & \cellcolor{my_blue}\textbf{30.68} & \cellcolor{my_blue}31.86 & \cellcolor{my_blue}\textbf{28.30} \\
\cmidrule[0.5pt](lr){2-5} \cmidrule[0.5pt](lr){6-9}
ControlNet
& 30K & 58.54 & 30.87 & 89.82
& 30K & 32.63 & 31.39 & 34.52 \\
\small{+ REPA}
& 18K & 62.23 & 30.81 & \textbf{92.75}
& 25K & 37.72 & 30.88 & 34.34 \\
\cellcolor{my_blue}\small\textbf{+ LISA (ours)}
& \cellcolor{my_blue}18K & \cellcolor{my_blue}\textbf{56.50} & \cellcolor{my_blue}\textbf{31.07} & \cellcolor{my_blue}89.90
& \cellcolor{my_blue}25K & \cellcolor{my_blue}\textbf{32.51} & \cellcolor{my_blue}\textbf{31.46} & \cellcolor{my_blue}\textbf{34.88} \\
\midrule
\multicolumn{1}{c}{\multirow{2}{*}{Method}}
& \multicolumn{4}{c}{Low-resolution Image} & \multicolumn{4}{c}{Depth Map} \\
\cmidrule(lr){2-5} \cmidrule(lr){6-9}
& Iter. & FID$_{\downarrow}$ & PSNR$_{\uparrow}$ & LPIPS$_{\downarrow}$
& Iter. & FID$_{\downarrow}$ & CLIP$_{\uparrow}$ & RMSE$_{\downarrow}$ \\
\midrule
T2I-Adapter
& \hphantom{0}5K & 31.76 & 16.17 & 0.457 & \hphantom{0}5K & 66.31 & 29.59 & 0.125 \\
\cellcolor{my_blue}\small\textbf{+ LISA (ours)}
& \cellcolor{my_blue}\hphantom{0}5K & \cellcolor{my_blue}\textbf{31.64} & \cellcolor{my_blue}\textbf{16.26} & \cellcolor{my_blue}\textbf{0.456} & \cellcolor{my_blue}\hphantom{0}5K & \cellcolor{my_blue}\textbf{62.70} & \cellcolor{my_blue}\textbf{29.77} & \cellcolor{my_blue}\textbf{0.121} \\
\cmidrule[0.5pt](lr){2-5} \cmidrule[0.5pt](lr){6-9}
ControlNet
& \hphantom{0}4K & 44.77 & 10.58 & 0.740
& \hphantom{0}2K & 67.68 & 28.10 & 0.162 \\
\cellcolor{my_blue}\small\textbf{+ LISA (ours)}
& \cellcolor{my_blue}\hphantom{0}4K & \cellcolor{my_blue}\textbf{34.47} & \cellcolor{my_blue}\textbf{15.48} & \cellcolor{my_blue}\textbf{0.506}
& \cellcolor{my_blue}\hphantom{0}2K & \cellcolor{my_blue}\textbf{65.23} & \cellcolor{my_blue}\textbf{28.13} & \cellcolor{my_blue}\textbf{0.153} \\
\cmidrule[0.5pt](lr){2-5} \cmidrule[0.5pt](lr){6-9}
ControlNet
& 20K & 23.96 & 16.89 & 0.426
& 10K & 77.84 & 25.61 & 0.120 \\
\small{+ REPA}
& 15K & 28.84 & 15.98 & 0.464
& \hphantom{0}4K & 72.65 & \textbf{27.29} & 0.122 \\
\cellcolor{my_blue}\small\textbf{+ LISA (ours)}
& \cellcolor{my_blue}15K & \cellcolor{my_blue}\textbf{23.87} & \cellcolor{my_blue}\textbf{17.51} & \cellcolor{my_blue}\textbf{0.425}
& \cellcolor{my_blue}\hphantom{0}4K & \cellcolor{my_blue}\textbf{66.75} & \cellcolor{my_blue}26.80 & \cellcolor{my_blue}\textbf{0.114} \\
\hline\hline
\end{tabularx}}
\vspace{-1em}
\end{table}

%% file: figures/fig_main.tex
\begin{figure}[t]
    \centering
    \includegraphics[width=1\linewidth]{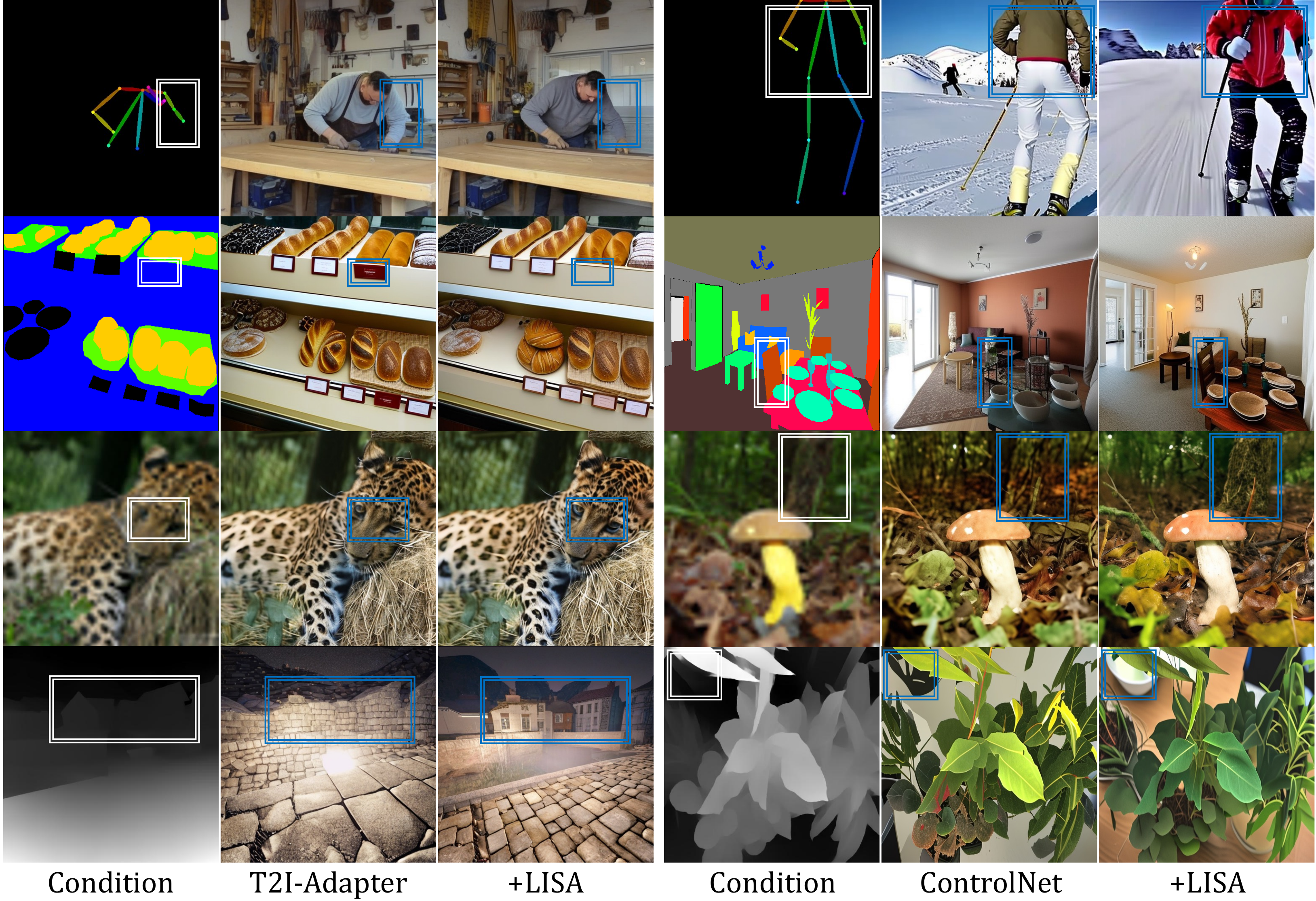}
    \vspace{-2em}
    \caption{\textbf{Qualitative examples across four image-condition generation tasks.} LISA
shows better condition-following performance (see highlighted parts in blue boxes).}
    \label{fig:main}
    \vspace{-1em}
\end{figure}

%% file: tables/table2.tex

\begin{wraptable}{r}{0.4\textwidth} 
\centering
\vspace{-2em}
\caption{\textbf{Ablation of alignment depth and $\lambda$}. The first row is the baseline.}
\label{tab:ablation}
\renewcommand{\arraystretch}{0.95} 
\resizebox{\linewidth}{!}{%
\begin{tabular}{ccccc}
\toprule
\hline
\multicolumn{1}{c}{Depth} & $\lambda$ & FID   & CLIP  & PCK(\%)   \\ \hline
- & -    & 56.71 & 31.24 & 85.97 \\
2     & 0.2    & 56.40 & 30.96 & 88.03 \\
8     & 0.2    & 56.75 & 31.14 & 88.06 \\
\cellcolor{my_blue}5     & \cellcolor{my_blue}0.2    & \cellcolor{my_blue}56.50 & \cellcolor{my_blue}31.07 & \cellcolor{my_blue}89.90 \\ 
5     & 0.5    & 56.34 & 31.08 & 87.83 \\
5     & 0.1    & 57.31 & 31.08 & 86.19 \\
\hline\bottomrule
\end{tabular}}
\vspace{-1em}
\end{wraptable}

%% file: tables/table3.tex

\begin{wraptable}{r}{0.3\textwidth} 
\centering
\vspace{-2em}
\caption{Computational overhead comparisons.}
\label{tab:overhead}
\renewcommand{\arraystretch}{1.2}
\setlength{\tabcolsep}{2.5pt}
\resizebox{\linewidth}{!}{%
\begin{tabular}{lccc}
\toprule\hline
           & \#Params & GPU & Time \\ \hline
ControlNet &  364.2M  & 21G & 2.1s \\
\cellcolor{my_blue}+LISA       &  \cellcolor{my_blue}364.6M  & \cellcolor{my_blue}21G & \cellcolor{my_blue}2.3s \\ \hline\bottomrule
\end{tabular}
}
\vspace{-1em}
\end{wraptable}

%% file: tables/table4.tex

\begin{table}[t]
    \centering
    \setlength{\tabcolsep}{5pt}
    \begin{minipage}[t]{0.45\columnwidth}
        \centering
        \caption{Compatible with Stable Diffusion 3.}
        \vspace{0pt}    
        \def\arraystretch{1.04}
        \resizebox{\linewidth}{!}{
        \begin{tabular}{lcccc}
        \hline\hline
        \multicolumn{1}{c}{Method} & Iter. & FID$_{\downarrow}$      & CLIP$_{\uparrow}$ & mIoU(\%)$_{\uparrow}$ \\\hline
        ControlNet                 & 1K    & 32.08 & \textbf{32.07}  & 20.81    \\
        \cellcolor{my_blue}\textbf{+ LISA}                       & \cellcolor{my_blue}1K    & \cellcolor{my_blue}\textbf{31.87}   & \cellcolor{my_blue}32.03 & \cellcolor{my_blue}\textbf{22.64}    \\ \cline{2-5} 
        ControlNet                 & 5K   & 31.69  & 32.02   & 26.01    \\
        \cellcolor{my_blue}\textbf{+ LISA}                       & \cellcolor{my_blue}5K   & \cellcolor{my_blue}\textbf{31.47}   & \cellcolor{my_blue}\textbf{32.03}  & \cellcolor{my_blue}\textbf{27.58}    \\ \hline\hline
        \end{tabular}
        }
        \label{tab:4}
    \end{minipage}%
    \hfill
    \setlength{\tabcolsep}{4.5pt}
    \begin{minipage}[t]{0.54\columnwidth}
        \centering
        \caption{Compatible with Video Generation.}
        \vspace{0pt}
        \def\arraystretch{1.06}
        \resizebox{\linewidth}{!}{
        \begin{tabular}{lccccc}
        \hline\hline
        \multicolumn{1}{c}{Method} & Iter. & FVD$_{\downarrow}$   & SSIM$_{\uparrow}$   & LPIPS$_{\downarrow}$ & PCK(\%)$_{\uparrow}$ \\\hline
        SVD Control                  & \hphantom{0}5K    & 10.57 & 0.852 & 19.07 & 30.22    \\
        \cellcolor{my_blue}\textbf{+ LISA}                       & \cellcolor{my_blue}\hphantom{0}5K    & \cellcolor{my_blue}\hphantom{0}\textbf{7.85}  & \cellcolor{my_blue}\textbf{0.882} & \cellcolor{my_blue}\textbf{10.62} & \cellcolor{my_blue}\textbf{57.00}    \\ \cline{2-6} 
        SVD Control                  & 30K   & \hphantom{0}7.79  & 0.902 & \hphantom{0}9.28  & 90.53    \\
        \cellcolor{my_blue}\textbf{+ LISA}                       & \cellcolor{my_blue}30K   & \cellcolor{my_blue}\hphantom{0}\textbf{6.89}  & \cellcolor{my_blue}\textbf{0.905} & \cellcolor{my_blue}\hphantom{0}\textbf{8.74}  & \cellcolor{my_blue}\textbf{91.86}    \\ \hline\hline
        \end{tabular}
        }
        \label{tab:5}
    \end{minipage}
    \vspace{-1em}
\end{table}


%% file: figures/fig_svd.tex
\begin{figure}[t]
    \centering
    \includegraphics[width=1\linewidth]{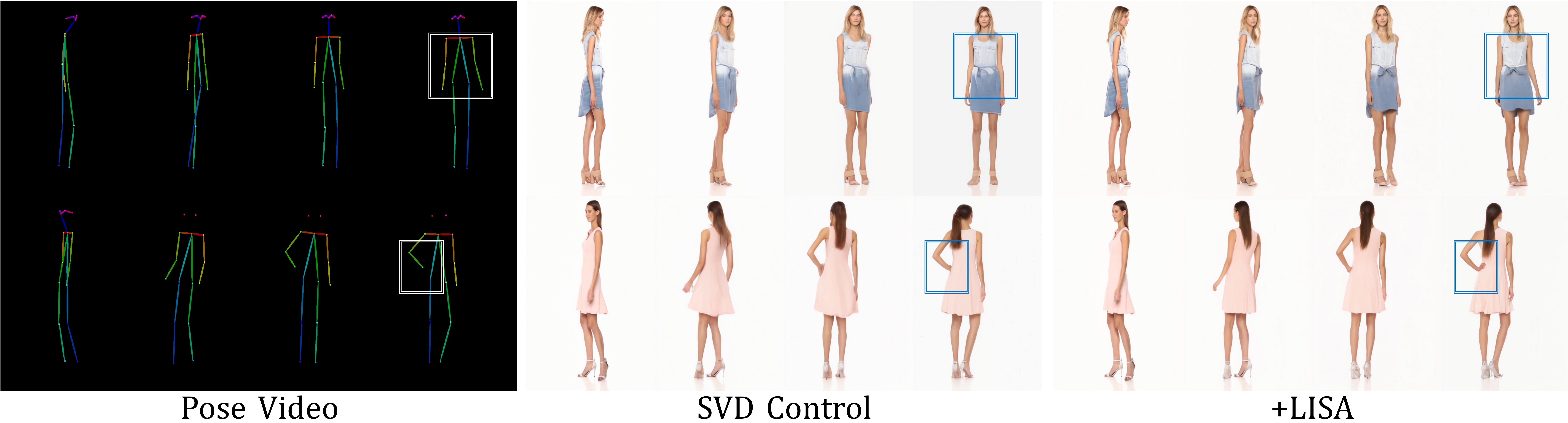}
    \vspace{-2em}
    \caption{\textbf{Qualitative comparisons on pose-conditioned video generation.} LISA shows better condition-following performance in the latter frame (see the highlighted parts in the blue boxes).}
    \vspace{-1em}
    \label{fig:svd}
\end{figure}

%% file: figures/fig_compose.tex
\begin{figure}[t]
    \centering
    \includegraphics[width=1\linewidth]{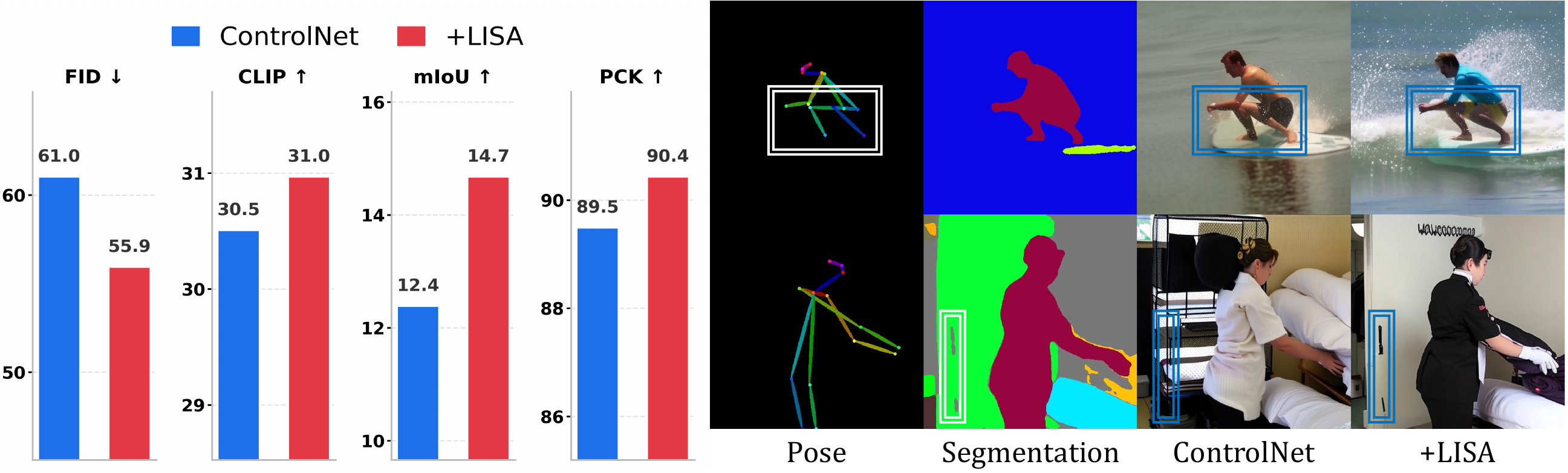}
    \vspace{-2em}
    \caption{\textbf{Quantitative (left) and qualitative (right) results of compositional-condition generation.} Benefiting from the explicit role decomposition, LISA shows better composition properties.}
    \label{fig:compose}
    \vspace{-1em}
\end{figure}

%% file: sections/5_related_works.tex
\section{RELATED WORK}
\label{sec:5}

\textbf{Conditional Visual Generation}. Besides the class and text~\citep{dhariwal2021diffusion,ho2022classifier}, the visual-modality conditions can provide spatial-structure guidance for generation. 
Composer~\citep{huang2023composer} trains a network that can adopt multi-modality conditions from scratch, thus can naturally achieve conditional control. However, the training cost limits its extension efficiency when facing new condition types. To this end, works~\citep{li2023gligen,zhang2023adding,mou2024t2i,strawberry_svdtemporalcontrolnet_2023,choi2025controllable} propose to freeze the pretrained diffusion model and fine-tune a side network (\eg, a copied encoder or condition adapter) and inject the condition feature into the original pretrained model. To further improve the controllability and efficiency, ControlNeXt~\citep{peng2024controlnext} aligns denoising distributions with the control features and ControlNet++~\citep{li2024controlnet++} incorporates reinforcement learning for post-training. Moreover, Uni-ControlNet~\citep{zhao2023uni} built a ControlNet that can model various visual conditions by training unified control adapters with extensive training data. Our method motivates the incorporation of an additional alignment to further improve efficiency.
 
\textbf{Training Diffusion Models with Regularizations}. The vanilla diffusion and flow matching models regress the target (\eg, noise, score, and velocity) as the main training loss. On top of that, some works propose adding an extra regularization during training to accelerate convergence. Studies~\citep{yu2024representation,pernias2024wurstchen,li2024return} leverage pretrained semantic visual encoders to help diffusion models' efficiency and final performance. $\Delta$FM~\citep{stoica2025contrastive} constructs a contrastive objective to regularize the flow trajectories and accelerate the training of the flow model. In video generation, works~\citep{wu2025geometry,huang2025jog3r,zhang2025endless} incorporate pretrained 3-D models' features or additional proxy 3-D tasks for training video diffusion models, enhancing consistency in synthetic videos. Our method shares a similar motivation with the above works, but differs in regularizing with the conditional probabilistic and score perspectives.

%% file: sections/6_conclusion.tex
\section{Conclusion}
\label{sec:6}

In this paper, we focus on the dual-branch paradigm for visual-condition controllable generation. Based on the role decomposition of its main and side networks from the score perspective, we propose LISA, which aligns the middle feature of the side network with a constructed likelihood score. By adding such a simple extra regularization objective, LISA can significantly accelerate training and bootstrap better synthetic results on perceptual quality and condition fidelity. Extensive ablations verify our consistent effectiveness and compatibility with U-Net/DiT architectures, diffusion/flow models, and image/video tasks. Besides, we found that LISA naturally shows better potential for compositional control, benefiting from the decomposition. In the future, we will extend the LISA regularization to practical applications and more general conditional generation scenarios.



%% file: sections/7_appendix.tex
\renewcommand{\thesection}{A.\arabic{section}}
\renewcommand*{\thesubsection}{A.\arabic{subsection}}
\renewcommand{\theequation}{A\arabic{equation}}
\renewcommand{\thetable}{A\arabic{table}}
\renewcommand{\thefigure}{A\arabic{figure}}

\newpage
\appendix
\noindent\textbf{\Large Appendix}
\vspace{0.1in}

This appendix is organized as follows:

\begin{itemize}[leftmargin=*]
\itemsep-0.2em
\item Section~\ref{sec:a.1} gives the implementation details of our method and reproduction details of the compared baselines.
\item Section~\ref{sec:a.2} provides the derivation of the compatibility of velocity-based Optimal Transport Flow Matching (OT-FM) and noise-based Variance-Preserving SDE (VP-SDE).
\item Section~\ref{sec:a.3} provides additional qualitative results.  
\end{itemize}

\newpage
\subsection{Implementation Details}
\label{sec:a.1}
We implement our method using PyTorch, Hugging Face Diffusers, and Accelerate, with Stable Diffusion 2.1-base (SD2.1) and Stable Video Diffusion (SVD) as the pretrained backbones. We freeze the pretrained denoising backbone, VAE, and text or image encoders, and jointly optimize ControlNet and a lightweight auxiliary decoder. The decoder receives intermediate ControlNet features before the zero-initialized output projection and predicts the residual between the denoising target and the frozen backbone prediction without control. It is used only during training and introduces no additional inference overhead. 

Unless otherwise specified, we extract features from \texttt{down\_5} (zero-based index) and set the auxiliary loss weight to $0.2$. All experiments are conducted on eight NVIDIA H20 GPUs. We optimize the trainable parameters using AdamW with a constant learning rate
of $10^{-5}$, $\beta_1=0.9$, $\beta_2=0.999$, and weight decay $0.01$. Training uses FP16 mixed precision, gradient checkpointing, and gradient clipping with a maximum norm of $1.0$. Image experiments use a resolution of $512\times512$, with a per-GPU batch size of 8 and four gradient accumulation steps, yielding an effective batch size of 256. 

For UBCFashion, we train SVD on 14-frame clips sampled at a frame interval of 4, with a resolution of $768\times512$ (height $\times$ width), a motion bucket ID of 127, and an effective batch size of 8. 

At inference, SD2.1 uses the DPM-Solver multistep scheduler with 20 sampling steps and a classifier-free guidance scale of $7.5$.

\newpage
\subsection{The Derivation of the Compatibility of OT-FM and VP-SDE}
\label{sec:a.2}

We developed our decomposition of the dual-branch paradigm from the score-based perspective in the main paper. This decomposition is also compatible with diffusion and flow matching models that predict noise and velocity. In this section, we provide the equivalent derivations for $v$-based Optimal Transport Flow Matching (OT-FM) and $\epsilon$-based Variance-Preserving SDE (VP-SDE).

Recall that the conditional score can be decomposed as
\begin{equation}
    \nabla_{\bm{x}_t}\log p_t(\bm{x}_t\mid\bm{c})
    = \nabla_{\bm{x}_t}\log p_t(\bm{x}_t)
    + \ell_t(\bm{c}\mid\bm{x}_t),
    \qquad
    \ell_t(\bm{c}\mid\bm{x}_t)
    := \nabla_{\bm{x}_t}\log p_t(\bm{c}\mid\bm{x}_t).
    \label{eq:lisa_param_bayes}
\end{equation}
Accordingly, the frozen main network provides $s_\theta(\bm{x}_t,t)$, while the side network induces the residual correction
\begin{equation}
    s_{\theta,\phi}(\bm{x}_t,\bm{c},t)-s_\theta(\bm{x}_t,t)
    \approx \ell_t(\bm{c}\mid\bm{x}_t).
    \label{eq:lisa_param_model_decomposition}
\end{equation}
This relation becomes exact when both predictors recover their corresponding population scores. We omit prediction arguments when the context is clear, and use the subscripts $\theta$ and $(\theta,\phi)$ for predictions without and with side-network condition injection, respectively.

For a Gaussian forward process $\bm{x}_t=\alpha_t\bm{x}_0+\sigma_t\bm{\epsilon}$, where $\bm{\epsilon}\sim\mathcal{N}(\bm{0},\mathbf{I})$ is independent of $(\bm{x}_0,\bm{c})$, the sample-wise approximated likelihood score used by LISA is
\begin{equation}
    \widehat{\ell}_t(\bm{c}\mid\bm{x}_t)
    = \nabla_{\bm{x}_t}\log p_t(\bm{x}_t\mid\bm{x}_0)-s_\theta
    = -\frac{\bm{x}_t-\alpha_t\bm{x}_0}{\sigma_t^2}-s_\theta.
    \label{eq:lisa_param_score_target}
\end{equation}
Its conditional expectation given $(\bm{x}_t,\bm{c})$ approximates $\ell_t(\bm{c}\mid\bm{x}_t)$, with equality for an exact main-network score. All conversions below are considered at interior times where $\alpha_t,\sigma_t>0$.

\subsubsection{OT-FM with $v$-prediction.}
In Flow Matching with this Gaussian path, the velocity field $v_\theta(\bm{x}_t,t)$ corresponds to the probability-flow ODE (PF-ODE) drift. Thus, we have the relations
\begin{equation}
    v_\theta=\bm{f}(\bm{x}_t,t)-\frac{1}{2}g^2(t)s_\theta,
    \qquad
    v_{\theta,\phi}=\bm{f}(\bm{x}_t,t)-\frac{1}{2}g^2(t)s_{\theta,\phi}.
    \label{eq:lisa_param_velocity_score}
\end{equation}
Substituting these relations into Eq.~(\ref{eq:lisa_param_model_decomposition}), we rewrite the decomposition in terms of velocity:
\begin{align}
    v_{\theta,\phi}-v_\theta
    &= -\frac{1}{2}g^2(t)\left(s_{\theta,\phi}-s_\theta\right) \nonumber\\
    &\approx -\frac{1}{2}g^2(t)\ell_t(\bm{c}\mid\bm{x}_t).
    \label{eq:lisa_param_velocity_residual}
\end{align}

For OT-FM, the trajectory is defined by $\sigma_t=1-\alpha_t$. Under a linear schedule where $\alpha_t=1-t$ and $\sigma_t=t$, we have $\dot{\alpha}_t=-1$ and $\dot{\sigma}_t=1$. This follows the data-to-noise time convention of the main paper; generation integrates from $t=1$ toward $t=0$. The corresponding linear SDE with the same Gaussian perturbation kernels has drift and squared diffusion coefficients
\begin{equation}
    \bm{f}(\bm{x}_t,t)
    =\frac{\dot{\alpha}_t}{\alpha_t}\bm{x}_t
    =-\frac{\bm{x}_t}{1-t},
    \label{eq:lisa_param_ot_drift}
\end{equation}
\begin{equation}
    g^2(t)
    =2\sigma_t\dot{\sigma}_t
    -2\frac{\dot{\alpha}_t}{\alpha_t}\sigma_t^2
    =-2\dot{\alpha}_t\frac{\sigma_t}{\alpha_t}
    =\frac{2t}{1-t}.
    \label{eq:lisa_param_ot_diffusion}
\end{equation}
Substituting these coefficients into Eq.~(\ref{eq:lisa_param_velocity_residual}) yields the velocity-based decomposition:
\begin{equation}
    v_{\theta,\phi}
    \approx v_\theta-\frac{t}{1-t}\ell_t(\bm{c}\mid\bm{x}_t).
    \label{eq:lisa_param_velocity_decomposition}
\end{equation}

We now express the alignment target in the same velocity space. The sample-wise velocity target along the linear trajectory is
\begin{equation}
    \bm{v}_t
    =\dot{\alpha}_t\bm{x}_0+\dot{\sigma}_t\bm{\epsilon}
    =\bm{\epsilon}-\bm{x}_0
    =\frac{\bm{x}_t-\bm{x}_0}{\sigma_t}.
    \label{eq:lisa_param_velocity_target}
\end{equation}
Multiplying Eq.~(\ref{eq:lisa_param_score_target}) by $-g^2(t)/2$ and using Eq.~(\ref{eq:lisa_param_velocity_score}), we obtain
\begin{align}
    -\frac{1}{2}g^2(t)\widehat{\ell}_t(\bm{c}\mid\bm{x}_t)
    &=\frac{g^2(t)}{2\sigma_t^2}
      \left(\bm{x}_t-\alpha_t\bm{x}_0\right)
      +\bm{f}(\bm{x}_t,t)-v_\theta \nonumber\\
    &=\frac{\bm{x}_t-\alpha_t\bm{x}_0}{\alpha_t\sigma_t}
      -\frac{\bm{x}_t}{\alpha_t}-v_\theta \nonumber\\
    &=\frac{\bm{x}_t-\bm{x}_0}{\sigma_t}-v_\theta
      =\bm{v}_t-v_\theta,
    \label{eq:lisa_param_velocity_alignment}
\end{align}
where the last line uses $1-\sigma_t=\alpha_t$. Consequently, the velocity residual $\bm{v}_t-v_\theta$ is the same approximated likelihood-score target scaled by $-t/(1-t)$.

\subsubsection{VP-SDE with $\epsilon$-prediction.}
For a VP forward process $\bm{x}_t=\alpha_t\bm{x}_0+\sigma_t\bm{\epsilon}$ constrained by $\alpha_t^2+\sigma_t^2=1$, the score function is parameterized as
\begin{equation}
    s_\theta=-\frac{\epsilon_\theta}{\sigma_t},
    \qquad
    s_{\theta,\phi}=-\frac{\epsilon_{\theta,\phi}}{\sigma_t}.
    \label{eq:lisa_param_noise_score}
\end{equation}
Substituting these relations into Eq.~(\ref{eq:lisa_param_model_decomposition}), we obtain
\begin{equation}
    -\frac{\epsilon_{\theta,\phi}-\epsilon_\theta}{\sigma_t}
    \approx\ell_t(\bm{c}\mid\bm{x}_t).
\end{equation}
This reveals the equivalent noise-based decomposition:
\begin{equation}
    \epsilon_{\theta,\phi}
    \approx\epsilon_\theta-\sigma_t\ell_t(\bm{c}\mid\bm{x}_t).
    \label{eq:lisa_param_noise_decomposition}
\end{equation}

Meanwhile, observing that $\bm{x}_t-\alpha_t\bm{x}_0=\sigma_t\bm{\epsilon}$, we can substitute Eq.~(\ref{eq:lisa_param_noise_score}) into the alignment target in Eq.~(\ref{eq:lisa_param_score_target}):
\begin{align}
    \widehat{\ell}_t(\bm{c}\mid\bm{x}_t)
    &=-\frac{\bm{\epsilon}}{\sigma_t}
      +\frac{\epsilon_\theta}{\sigma_t} \nonumber\\
    &=-\frac{\bm{\epsilon}-\epsilon_\theta}{\sigma_t}.
    \label{eq:lisa_param_noise_alignment}
\end{align}
Thus, the corresponding alignment target in noise space is simply $\bm{\epsilon}-\epsilon_\theta$, which is the approximated likelihood score multiplied by $-\sigma_t$.

\paragraph{Equivalent alignment across parameterizations.}
These derivations show that the side network induces a correction proportional to the likelihood score in each prediction space. Its alignment target can be constructed as the standard training target minus the frozen main-network prediction. To preserve the score-space LISA objective exactly, the same conversion must also be applied to the decoder output and the loss weight. Specifically, let $a_t=-t/(1-t)$ or $-\sigma_t$ for the OT-FM velocity or VP noise parameterization, respectively. Writing the converted decoder output as $\mathcal{D}^{(a)}_\psi=a_t\mathcal{D}_\psi$, we have
\begin{equation}
    \mathcal{L}_{\mathrm{LISA}}
    =\mathbb{E}_{\bm{x}_0,\bm{c},t,\bm{x}_t}\left[
    \frac{1}{a_t^2}
    \left\|\mathcal{D}^{(a)}_\psi
    -\operatorname{sg}\!\left[a_t\widehat{\ell}_t(\bm{c}\mid\bm{x}_t)\right]
    \right\|_2^2\right],
    \label{eq:lisa_param_alignment_loss}
\end{equation}
where the decoder arguments are omitted for brevity and $\operatorname{sg}$ denotes stop-gradient. Omitting the factor $1/a_t^2$ instead gives a time-reweighted alignment objective; the underlying decomposition and target conversion remain the same.

\newpage
\subsection{Additional Results}
\label{sec:a.3}
We provide the visualization results of different diffusion backbones in Figure~\ref{fig:sd3} and more qualitative comparisons with ControlNet in Figure~\ref{fig:app}.

\input{figures/fig_sd3}
\input{figures/fig_appendix}

%% file: figures/fig_sd3.tex
\begin{figure}[h]
    \centering
    \includegraphics[width=1\linewidth]{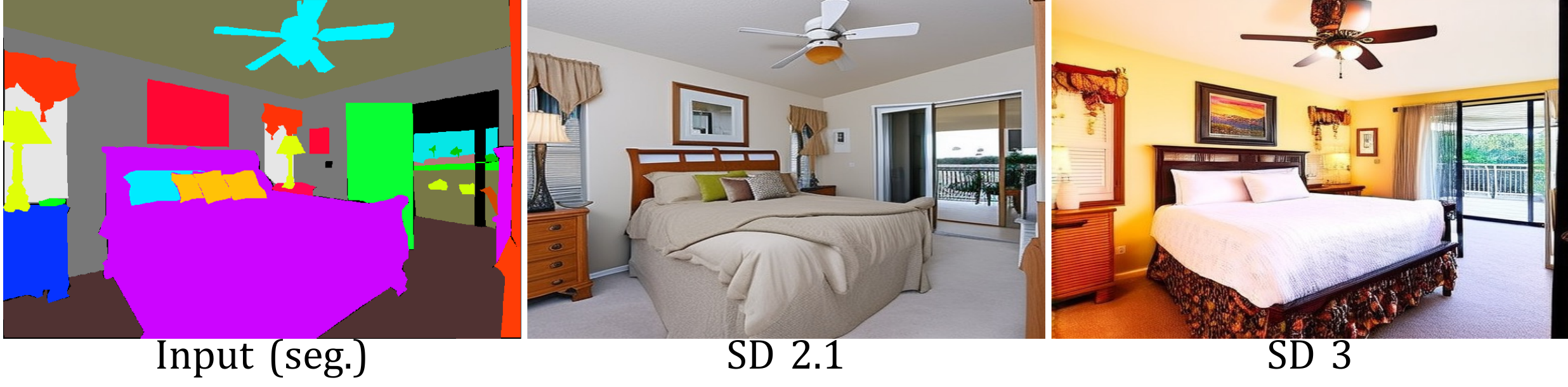}
    \caption{Visualization examples across SD2.1 and SD3.}
    \label{fig:sd3}
\end{figure}

%% file: figures/fig_appendix.tex
\begin{figure}[h]
    \centering
    \includegraphics[width=1\linewidth]{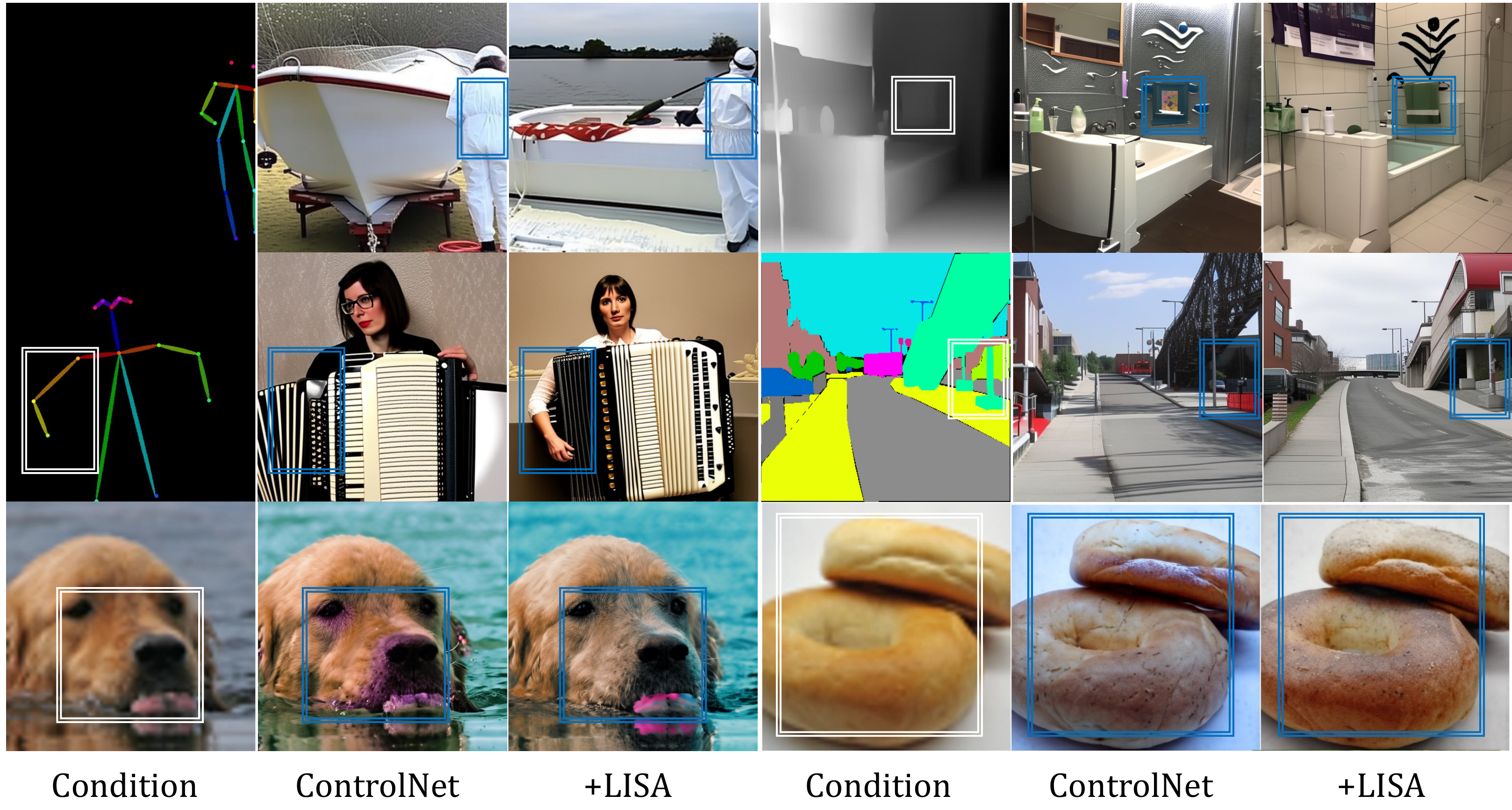}
    \caption{Additional qualitative comparisons with ControlNet across four image-condition generation tasks.}
    \label{fig:app}
\end{figure}